\documentclass[letterpaper, 10 pt, conference]{ieeeconf}  

\IEEEoverridecommandlockouts                              

\usepackage{tcolorbox}
\usepackage{amsmath}

\usepackage{hyperref}
\usepackage{cleveref}
\usepackage{algorithm}
\usepackage{algorithmic}
\usepackage{amssymb}
\usepackage{booktabs}
\usepackage{multirow}
\usepackage[table]{xcolor}
\usepackage{graphicx}
\usepackage{capt-of}
\usepackage{booktabs}
\usepackage{makecell}

\title{\LARGE \bf
Learning Safe-Stoppability Monitors for Humanoid Robots
}

\author{
Yifan Sun$^{1}$, Yiyuan Pan$^{1}$, Shangtao Li$^{1}$, 
Caiwu Ding$^{2}$, Tao Cui$^{2}$, Lingyun Wang$^{2}$, 
Changliu Liu$^{1}$\\
$^{1}$Robotics Institute, Carnegie Mellon University, Pittsburgh, PA, USA\\
$^{2}$Foundational Technologies, Siemens Corporation, Princeton, NJ, USA\\
\texttt{\{yifansu2, yiyuanp, shangtal, cliu6\}@andrew.cmu.edu}\\
\texttt{\{caiwu.ding, tao.cui, max.wang\}@siemens.com}
}

\begin{document}

\makeatletter
\let\@oldmaketitle\@maketitle
    \renewcommand{\@maketitle}{\@oldmaketitle
    \centering
    \includegraphics[width=1.0\textwidth]{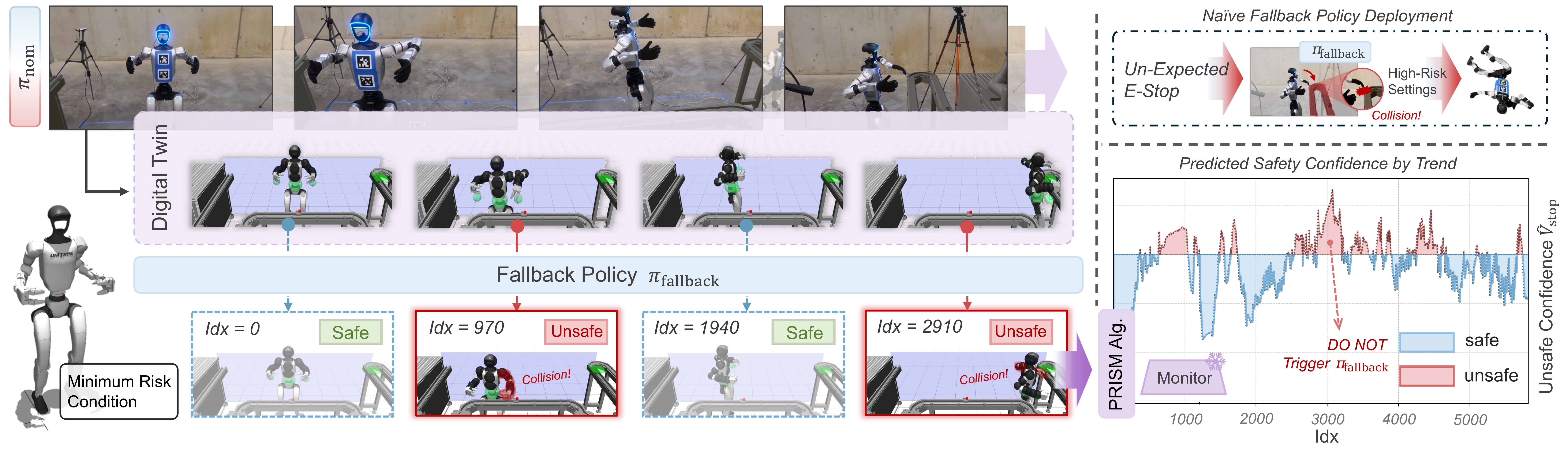}
    \captionof{figure}{
    \textbf{Safe-stoppability monitoring for humanoid robots.} Emergency stops for humanoids cannot simply cut power; instead, a predefined fallback controller is triggered to drive the robot toward a minimum-risk condition (MRC). We simulate fallback executions from nominal states in a digital twin to label whether the robot can safely reach the terminal configuration without collision or failure. These labels are used to train a neural monitor that predicts the real-time safe-stoppability confidence of the current state. The monitor can not only discourage pre-mature execution of the fallback policy, but also enable proactive intervention by triggering the fallback policy before the robot enters states from which safe stopping is no longer possible. 
    }
    \label{fig:firstpage}
    \setcounter{figure}{1}
  }
\makeatother

\maketitle
\thispagestyle{empty}
\pagestyle{empty}

\begin{abstract}
Emergency stop (E-stop) mechanisms are the de facto standard for robot safety. However, for humanoid robots, abruptly cutting power can itself cause catastrophic failures; instead, an emergency stop must execute a predefined fallback controller that preserves balance and drives the robot toward a minimum-risk condition. This raises a critical question: from which states can a humanoid robot safely execute such a stop?
In this work, we formalize emergency stopping for humanoids as a policy-dependent safe-stoppability problem and use data-driven approaches to characterize the safe-stoppable envelop. 
We introduce \textbf{PRISM} (\textbf{P}roactive \textbf{R}efinement of \textbf{I}mportance-sampled \textbf{S}toppability \textbf{M}onitor), a simulation-driven framework that learns a neural predictor for state-level stoppability. 
PRISM iteratively refines the decision boundary using importance sampling, enabling targeted exploration of rare but safety-critical states.
This targeted exploration significantly improves data efficiency while reducing false-safe predictions under a fixed simulation budget.
We further demonstrate sim-to-real transfer by deploying the pretrained monitor on a real humanoid platform. 
Results show that modeling safety as policy-dependent stoppability enables proactive safety monitoring and supports scalable certification of fail-safe behaviors for humanoid robots. The project website can be found at 
\href{https://intelligent-control-lab.github.io/humanoid_stoppability/}{\nolinkurl{https://intelligent-control-lab.github.io/humanoid_stoppability/}}.

\end{abstract}

\section{INTRODUCTION}

Humanoid robots are increasingly deployed in unstructured, human-centered environments where unexpected contacts, perception errors, model mismatch, or external disturbances may arise during task execution. In such settings, a fundamental safety layer is the emergency-stop (E-stop) mechanism. Unlike fixed-base industrial manipulators, however, a humanoid robot cannot simply cut power when an E-stop is triggered. Due to their underactuated nature and reliance on unilateral ground contacts, abrupt power loss inevitably induces catastrophic falls and secondary collisions. Instead, humanoid E-stop must execute a predefined fallback controller that preserves balance, regulates contacts, and drives the robot toward a safe terminal configuration.

On the other hand, E-stop triggers are asynchronous and unpredictable. An emergency stop may be triggered at any time due to human intrusion, system-level anomalies, communication failures, or direct operator intervention. Given the temporal unpredictability of these interventions, the robot must maintain its state in the \textit{safe-stoppable envelope (SSE)}: a set of states from which the fallback controller can reliably drive the robot to a dynamically stable terminal condition, usually called \textit{minimum risk condition} (MRC).
This requirement leads to a state-dependent fail-safe property that we call \textit{safe-stoppablity}. Rather than requiring invariant safety under nominal control, safe-stoppability asks a more urgent question: 

\begin{tcolorbox}[top=1pt, bottom=1pt, left=0pt, right=0pt]
\begin{center}
\textit{If we press the button, would the robot survive the stop?}
\end{center}
\end{tcolorbox}


We formalize this property as a stochastic reach-avoid condition under the fixed fallback policy. A state is \textit{safely stoppable} if, when the fallback controller is triggered, the robot can reach MRC while avoiding intermediate failures such as falls, forbidden collisions, or joint-limit violations. 
Importantly, this definition naturally excludes states from which failure is inevitable before the fallback policy can bring the system to the MRC, as such states lie outside SSE. 

This perspective leads to a critical insight: a safe-stoppability monitor can serve not only as a \textit{passive} diagnostic tool after an external E-stop trigger, but also as a \textit{proactive} supervisory mechanism. By continuously evaluating whether the current state remains within SSE, the monitor can autonomously trigger E-stop behavior before catastrophic failure occurs, effectively preventing falls or collisions that would otherwise unfold under nominal control. In this way, safe-stoppability becomes a runtime assurance mechanism that bridges nominal task execution and emergency intervention.

Nevertheless, precisely characterizing SSE for high-dimensional humanoids is fundamentally challenging. First, exact reach–avoid computation under nonlinear, contact-rich hybrid dynamics is analytically intractable and numerically prohibitive. Second, directly learning the boundary from hardware data is unsafe and impractical, as failure trials are costly and potentially catastrophic. Third, failure-to-stop events lie in the long tail of the state distribution, making naive data collection highly inefficient and poorly conditioned near the boundary of SSE between stoppable and non-stoppable states. 

To address these challenges, we introduce \textbf{PRISM} (\textbf{P}roactive \textbf{R}efinement of \textbf{I}mportance-sampled \textbf{S}toppability \textbf{M}onitors), a simulation-driven framework that learns a neural safe-stoppability monitor under a predefined fallback controller for E-stop. The predefined fallback controller drives the humanoid to a MRC, defined as a stable standing posture with both feet firmly grounded and both arms raised alongside the body. This configuration provides a general-purpose, dynamically stable posture that minimizes risk during emergency termination. 
On the other hand, since safety behaviors are highly context dependent, we restrict our scope to a known task and environment, focusing on the nominal state distribution induced during execution rather than the entire state space. 
Whether the state is inside the SSE is labeled in large-scale simulation, and an iterative importance-guided refinement strategy concentrates data collection near the predicted boundary between stoppable and non-stoppable states. By strictly targeting these high-uncertainty regions rather than exhaustive exploration, PRISM yields a highly data-efficient demarcation of SSE, enabling reliable sim-to-real deployment.
The main contributions of this work are:

\begin{itemize}
    \item \textbf{Characterization of Safe-stoppability.} We formalize fail-safe safety for humanoid robots as a policy-dependent stochastic reach-avoid property under a fixed fallback controller for E-stop, and propose a data-driven method to learn the neural safe-stoppability monitor.

    \item \textbf{Simulation-driven boundary refinement.} We propose an iterative, importance-guided sampling framework that dynamically reallocates sampling density toward informative boundary regions, improving boundary characterization for the safe-stoppable envelop (SSE).


    \item \textbf{Sim-to-real validation.} We demonstrate real-world deployment on a Unitree G1 humanoid. Our framework enables efficient and safe verification of predefined fallback policies, making the characterization of the safe-stoppable envelop (SSE) on hardware significantly more practical and scalable.

\end{itemize}

By establishing verifiable safe-stop capabilities for bipedal humanoid robots, this work tackles one of the most significant obstacles preventing their adoption in manufacturing environments; demonstrating fail-safe operation that meets stringent industrial safety requirements.

\section{Related Work}
\label{sec:related_work}

\begin{figure*}[t]
\centering
\includegraphics[width=\linewidth]{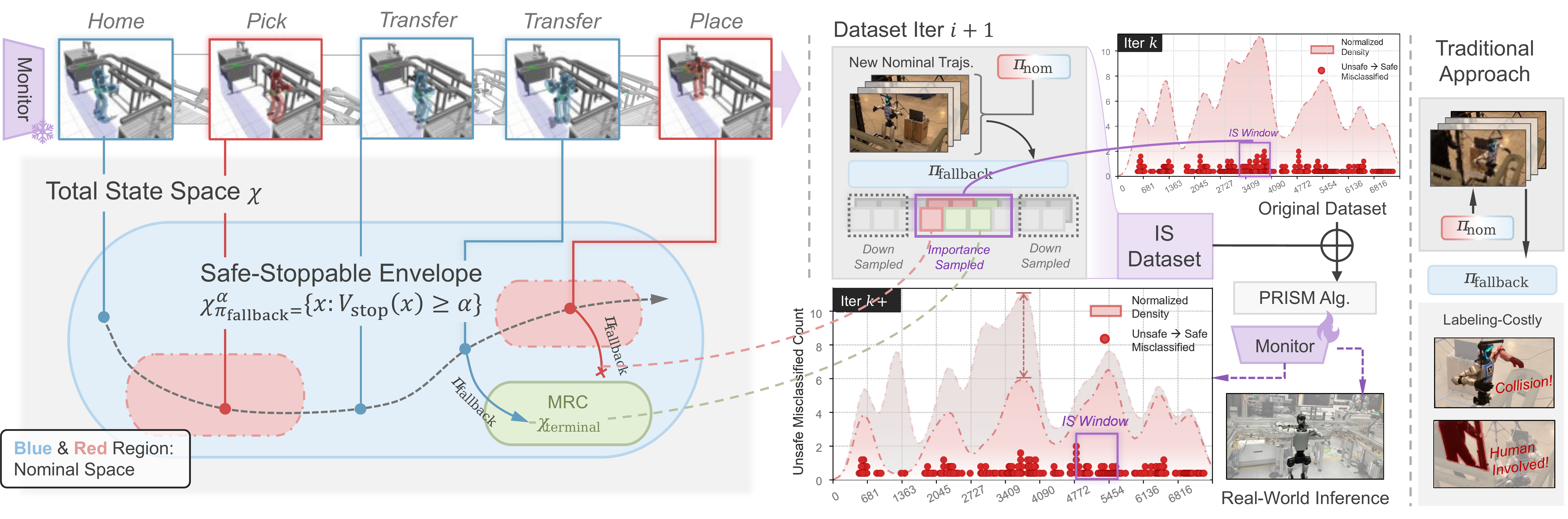}
\caption{\textbf{Data-efficient stoppability monitoring via PRISM.} Exhaustively collecting failure data to learn SSE boundaries on physical humanoids is prohibitively expensive and risks severe hardware damage. To overcome this labeling bottleneck, PRISM dynamically tracks prediction errors and allocates dense sampling to highly uncertain boundary regions (the IS Window). By focusing on critical unstoppable states rather than trivially safe regions, PRISM reduces data requirements while improving boundary characterization. Red dots denote counts, while the shaded background indicates density values.}
\label{fig:concept_and_sampling}
\end{figure*}

\subsection{Fail-Safe Safety and Minimum Risk Conditions}

Fail-safe operation is a foundational principle in safety-critical systems. Upon fault detection, systems are not expected to simply disable actuation; instead, they must transition to a configuration with bounded and acceptable residual risk (e.g., an autonomous vehicle safely pulling over to the roadside). 
In automotive safety, this requirement is formalized through the concept of \emph{Minimum Risk Condition} (MRC), defined in standards such as ISO~26262\cite{ISO26262_2018} and \emph{Safety Of The Intended Functionality} (SOTIF) defined by ISO~21448\cite{iso21448_2022}. 
A similar principle should be applied in humanoid robotics. 

On the other hand, prior work in humanoid control has extensively addressed disturbance rejection and balance recovery, including capture-point methods \cite{pratt2006capture}, push recovery \cite{stephens2011push}, step timing adjustment \cite{griffin2017walking}, and optimization-based whole-body control for platforms such as Atlas \cite{feng2014optimization}. More recent efforts explore multi-constraint safe control \cite{chen2025dexterous} and safety benchmarking toolkits \cite{sun2025spark}.
While these approaches improve robustness and disturbance handling, they do not explicitly characterize the state-dependent region from which a predefined emergency behavior can safely drive the robot to a minimum-risk terminal condition.



\subsection{Reach-Avoid Analysis for Fail-Safe Control}

Safety in control theory is often formulated as a \textit{invariance} problem, where the objective is to keep system trajectories within a safe set using supervisory control. Classical set-theoretic approaches characterize maximal controlled invariant sets for constrained systems \cite{blanchini1999set, liu2014control,ames2019control}. 

Fail-safe behavior, however, is inherently a \textit{reachability} problem: when safety cannot be guaranteed indefinitely, the system must reach a designated safe terminal condition while avoiding failure along the way. Reachability and reach-avoid analysis characterize states that can reach a target set while avoiding unsafe regions \cite{mitchell2005time, fisac2015reach}, with stochastic reachability extending these formulations to uncertain systems \cite{soudjani2013probabilistic, prandini2006stochastic}. While these methods provide strong theoretical guarantees, their computational complexity scales poorly with system dimension, making exact computation impractical for high-dimensional humanoid models.


Recent work has therefore explored data-driven approximations of reach-avoid sets \cite{fisac2018general, nakamura2025generalizing, pandya2025robots}. In many such approaches, a fixed policy is evaluated by estimating the probability of reaching a goal while avoiding failure, yielding a policy-dependent reach-avoid value function. Our formulation adopts this perspective but introduces iterative importance sampling to concentrate data collection near the predicted stoppability boundary, significantly improving data efficiency.

\subsection{Sim-to-Real Deployment for Humanoid Robots}

Simulation has become a standard tool for improving data efficiency in humanoid control. Large-scale parallel simulators~\cite{makoviychuk2021isaac} enable extensive exploration of robot dynamics and policy behavior without the cost and risk of real-world experimentation. Such simulation-driven pipelines have been widely adopted in humanoid locomotion and whole-body control~\cite{liao2025beyondmimic, he2025asap, ze2025twist2}, where policies are trained or evaluated using large numbers of simulated rollouts before deployment on physical systems. 
However, simulation-based methods must address the sim-to-real gap, arising from mismatches in friction, actuator dynamics, compliance, latency, and other physical effects. A common mitigation strategy is domain randomization, where physical parameters are randomized during simulation to encourage policies that generalize to real-world dynamics~\cite{tobin2017domain, peng2018sim}.

In this work, we adopt simulation to efficiently approximate the SSE and train the stoppability monitor. To reduce sim-to-real mismatch, simulation rollouts are initialized using real-robot state logs before executing the fallback controller. This design eliminates discrepancies along the nominal trajectory and confines the remaining gap to fallback execution. The residual mismatch is then mitigated through targeted domain randomization, improving the reliability of the learned monitor during real-world deployment.

\section{Problem Formulation}

This paper studies whether a predefined fallback (E-stop) policy can be safely executed at states in the \emph{nominal state space} induced by a task policy $\pi_{\text{nom}}$. 

\subsection{System Model}

Consider a humanoid robot with state $x_t \in \mathcal{X}$ and control input $u_t \in \mathcal{U}$ evolving under stochastic dynamics
$
x_{t+1} = f(x_t, u_t, w_t),
$
where $w_t$ captures disturbances, contact uncertainty, and modeling mismatch.

For a given task, the robot executes a nominal policy $u_t = \pi_{\text{nom}}(x_t)$, inducing a nominal state distribution $\mathcal{X}_{\text{nom}} \subset \mathcal{X}$.

We define a safe set $\mathcal{X}_{\text{safe}} \subset \mathcal{X}$ encoding admissible conditions such as no falling and no forbidden collisions. A predefined fallback (E-stop) policy
$
u_t = \pi_{\text{fallback}}(x_t)
$
is designed to drive the robot toward MRC characterized as a terminal stopping pose 
$
\mathcal{X}_{\text{terminal}} \subset \mathcal{X}_{\text{safe}}
$
as shown in \cref{fig:firstpage}. 

It is assumed that both the nominal policy and the fallback policy are well trained with high success rates if we do not account for complex environmental interaction.

\subsection{Safe-Stoppable Envelop (SSE) and Stoppability Monitor}

Due to stochastic disturbances, the success of a safe stop is inherently probabilistic. We therefore define the \emph{stoppability value function} using a reach–avoid problem formulation.
\begin{equation}
\begin{aligned}
V_{\text{stop}}(x_{\rm trig})
= \Pr\!\Big(
& \exists T \le T_{\max} \ \text{s.t.} \ 
x_T \in \mathcal{X}_{\text{terminal}}, \\
& x_t \in \mathcal{X}_{\text{safe}} \ \forall t \in [0,T]
\ \Big| \ x_{\rm trig}
\Big),
\end{aligned}
\end{equation}
which represents the probability that the fallback policy successfully reaches the terminal region while avoiding intermediate safety violations, given an instantaneous trigger state $x_{\rm trig}$ sampled during nominal execution.

For a confidence threshold $\alpha \in (0,1]$, the corresponding $\alpha$-safe-stoppable set is defined as the level set
\begin{equation}
\mathcal{X}^{\alpha}_{\pi_{\text{fallback}}}
=
\{ x \in \mathcal{X} \mid V_{\text{stop}}(x) \ge \alpha \}.
\end{equation}

To map the probabilistic outcome $V_{\text{stop}}(x)$ to a deterministic binary classification, we define a state $x \in \mathcal{X}$ as \textit{Stoppable} if $V_{\text{stop}}(x) \ge \alpha$, meaning the fallback policy succeeds with probability at least $\alpha$. Conversely, a state is considered \textit{Unstoppable} if $V_{\text{stop}}(x) < \alpha$. The SSE is then directly equivalent to the $\alpha$-level set $\mathcal{X}^{\alpha}_{\pi_{\text{fallback}}}$. By adjusting the threshold $\alpha$, we can recover different levels of robustness for the SSE, effectively controlling the trade-off between safety conservatism and nominal task efficiency.

The true stoppability function $V_{\text{stop}}(x)$ and SSE are unknown. 
Our objective is to learn an online neural monitor $\hat{V}_{\text{stop}}(x)$ to approximate the true stoppability $V_{\text{stop}}(x)$. 
Specifically, we aim to train this monitor such that:
\begin{enumerate}
    \item $\hat{V}_{\text{stop}}(x)$ accurately estimates the safe-stop success probability over the nominal state distribution $\mathcal{X}_{\pi_{\text{nom}}}$.
    \item The monitor maintains high reliability in rare but safety-critical regions, particularly reducing false-safe predictions (unstoppable states classified as stoppable).
\end{enumerate}

At runtime, the monitor triggers a proactive stop whenever $\hat{V}_{\text{stop}}(x) < \alpha$. 
The threshold $\alpha$ acts as a tunable safety margin that determines the required confidence level. In practice, $\alpha$ can be selected empirically to control the false-safe rate.
Even within the nominal state space, two key challenges arise when learning $\hat{V}_{\text{stop}}(x)$.

\textit{First}, stoppable and unstoppable states are often highly imbalanced, so uniform data sampling in $\mathcal{X}_{\text{nom}}$ wastes most samples on easy nominal states and fails to adequately cover the rare boundary states that are most informative for learning.

\textit{Second}, stoppability labels are expensive to obtain. Unlike standard safety events such as collisions or falls, which can be checked instantaneously, stoppability must be determined by executing the full fallback policy from a given state. On physical hardware, this is especially inefficient because the robot cannot generally be reset directly to arbitrary intermediate states, so each queried state must first be reached through rollout.

To address \textit{imbalance}, we develop the iterative importance sampling scheme in \Cref{sec:importance_sampling}, which focuses data collection on informative states using the monitor learned in the previous iteration. To address \textit{labeling cost}, we use simulation as a state replay engine in \Cref{sec:stoppability_estimation}, allowing direct initialization at queried states before fallback execution. The former improves where data are sampled; the latter reduces the cost of labeling each sampled state.

\section{Simulation-Based Stoppability Estimation}
\label{sec:stoppability_estimation}
\subsection{Efficient Safe-Stop Outcome Labeling}

We generate binary supervisory labels via simulated rollouts of $\pi_{\text{fallback}}$. Specifically, we sample instantaneous trigger states, denoted as $x_{\text{trig}}$, from a nominal trajectory distribution $q(x)$ (initially approximating $\mathcal{X}_{\pi_{\text{nom}}}$), execute the fallback policy $\pi_{\text{fallback}}$ for a maximum horizon $T_{\max}$, and assign the label.
\begin{equation}
y =
\begin{cases}
1, & \exists T \le T_{\max} \ \text{s.t.} \ 
x_T \in \mathcal{X}_{\text{terminal}}, \\
& x_t \in \mathcal{X}_{\text{safe}} \ \forall t \in [0,T],\ x_0 = x_{\text{trig}}, \\
0, & \text{otherwise.}
\end{cases}
\end{equation}


Fundamentally, the outcome of any single fallback rollout is a stochastic realization $y \in \{0, 1\}$, drawn from a Bernoulli distribution parameterized by the true safety probability: $y \sim \text{Bernoulli}(V_{\text{stop}}(x))$. Although each training sample provides only a deterministic binary label, optimizing the neural network via a binary cross-entropy objective drives the model to recover the conditional expectation $\mathbb{E}[y|x]$. Consequently, the optimal predictor converges precisely to the true probability $\Pr(y=1 | x) = V_{\text{stop}}(x)$. This statistical equivalence mathematically bridges the conceptual gap between single-rollout deterministic labels and stochastic reachability. 

\subsection{Learning the Stoppability Monitor}
Given the dataset $\mathcal{D}=\{(x^i_t,y^i_t)\}_{t=0,i=1}^{t=T_i,i=N}$, we train the $\theta$-parameterized neural network monitor $\hat{V}_{\text{stop}}(x;\theta)$ by minimizing a weighted binary cross-entropy loss
\begin{equation}
\ell(x,y) := \min_{\theta} \ \sum_{i=1}^N\sum_{t\in T^i_s} \, \omega(y_t^i)\,\ell\!\left(\hat{V}_{\text{stop}}(x_t^i;\theta), y_t^i\right). \label{loss_eq}
\end{equation}
where $\omega(\cdot)$ addresses class imbalance and $\ell(\cdot,\cdot)$ is the cross-entropy loss. $T_s^i$ denote the specific subset of sampled time steps for the $i$-th trajectory.
At runtime, the monitor produces a scalar estimate $\hat{V}_{\text{stop}}(x)$ and a binary stoppability decision:
\begin{equation}
\hat{x}\in \hat{\mathcal{X}}^{\alpha}_{\pi_{\text{fallback}}}
\iff
\hat{V}_{\text{stop}}(x) \ge \alpha.
\end{equation}

\section{Iterative Refinement via Importance Sampling}
\label{sec:importance_sampling}

During robust nominal execution, catastrophic failures under the fallback policy are inherently rare, as the system state predominantly resides within SSE. Uniformly sampling states $x \in \mathcal{X}_{\text{nom}}$ therefore produces predominantly successful rollouts, yielding limited information about the boundary of SSE. This severe class imbalance makes accurate learning of a stoppability monitor particularly challenging.

A naïve strategy would continually augment the dataset with additional nominal trajectories and corresponding safe-stop labels. However, this approach is highly data-inefficient: most nominal states are trivially stoppable and contribute little new information, while only a small subset of boundary or contact-critical states are informative. 

To overcome these limitations, we adopt a progressive refinement strategy that simultaneously (i) expands nominal state coverage across iterations and (ii) concentrates labeling effort on regions where the current monitor exhibits systematic error. Instead of balancing data post hoc, PRISM performs importance-guided resampling that prioritizes SSE boundary-proximal or high-uncertainty states. Because identifying such states requires a trained model, the framework alternates between monitor training and targeted resampling, yielding an iterative boundary-refinement procedure.

\subsection{Initial Dataset Construction}

We begin by collecting $N_0$ nominal trajectories under the nominal task policy $\pi_{\text{nom}}$, inducing a state distribution $p_{\text{nom}}^{(0)}(x)$. States are uniformly sampled along the time dimension and labeled via e-stop rollouts under $\pi_{\text{fallback}}$, yielding an initial dataset $\mathcal{D}^{(0)} = \{(x_t^i, y_t^i)\}_{t=0, i=1}^{t=T_i, i=N-0}$.
We train the initial stoppability monitor $\hat{V}_{\text{stop}}^{(0)}$ on $\mathcal{D}^{(0)}$.

\subsection{Data Refinement via Importance Sampling}

At iteration $k$, we generate a new batch of nominal trajectories (fixed number $N_I$), inducing distribution $p_{\text{nom}}^{(k)}(x)$. To identify informative regions, we maintain a fixed validation set
$\mathcal{D}_{\text{val}} = \{(x_t^i, y_t^ii)\}_{t=0,i=1}^{t=T_i,i=N_{\text{val}}}$ to compute the residuals of the current monitor:
$r_i^{(k)} =
\ell\!\left(x^i,\, y^i\right), (x^i, y^i) \in\mathcal{D}_{\text{val}}$ (for simplicity, we use $x^i,y^i$ to denote the $i$-th trajectory and its corresponding labels; see \cref{loss_eq}).
Let $q_{1-\delta}^{(k)}$ denote the empirical $(1-\delta)$ quantile of $\{r_i^{(k)}\}$.  
Inspired by conformal prediction~\cite{shafer2008tutorial,pan2025seeing}, this quantile provides a finite-sample calibration threshold to define the importance region as:
\begin{equation}
\mathcal{T}^{(k)} 
=
\left\{
x :
\ell\!\left(\hat{V}_{\text{stop}}^{(k)}(x),\, y(x)\right)
\ge
q_{1-\delta}^{(k)}
\right\},
\end{equation}

Intuitively, $\mathcal{T}^{(k)}$ captures states where the monitor’s prediction error exceeds the calibrated tolerance level, focusing refinement on systematically mispredicted regions.

\begin{algorithm}[t]
\caption{PRISM: \textbf{P}roactive \textbf{R}efinement of \textbf{I}mportance-sampled \textbf{S}toppability \textbf{M}onitors}
\label{alg:progressive_refine}
\begin{algorithmic}[1]
\STATE Collect $N_0$ nominal trajectories under $\pi_{\text{nom}}$ from initial state set $\mathcal{X}_{0}$.
\STATE Uniformly sample states and label via $\pi_{\text{fallback}}$ rollouts.
\STATE Train initial monitor $\hat{V}_{\text{stop}}^{(0)}$.
\STATE Initialize dataset $\mathcal{D}^{(0)}$ and validation set $\mathcal{D}_{\text{val}}$.

\FOR{$k = 0,1,\dots,K-1$}
    \STATE Generate new nominal trajectories with $\pi_{\text{nom}}$ from $\mathcal{X}_{0}$ inducing $p_{\text{nom}}^{(k)}(x)$.
    
    \STATE Compute residuals on validation set:
    \[
    r_i^{(k)} = 
    \ell\!\left(\hat{V}_{\text{stop}}^{(k)}(x_i), y_i\right), (x_i, y_i) \in \mathcal{D}_{\text{val}}.
    \]
    
    \STATE Compute quantile threshold $q_{1-\delta}^{(k)}$.
    
    \STATE Define importance region $\mathcal{T}^{(k)}$.
    
    \STATE Construct sampling distribution:
    \[
    p_\text{{mix}}^{(k)}(x)
    =
    (1-\beta)p_{\text{nom}}^{(k)}(x)
    +
    \beta p_{\text{imp}}^{(k)}(x).
    \]
    
    \STATE Sample states $x \sim p_\text{{mix}}^{(k)}(x)$.
    
    \FOR{each sampled state $x$}
        \STATE Roll out safe-stop under $\pi_{\text{fallback}}$ and obtain label $y$.
        \STATE Add $(x,y)$ to $\mathcal{D}^{(k+1)}$.
    \ENDFOR
    
    \STATE Retrain monitor $\hat{V}_{\text{stop}}^{(k+1)}$ with $\mathcal{D}^{(k+1)}$.
\ENDFOR

\STATE \textbf{return} $\hat{V}_{\text{stop}}^{(K)}$.
\end{algorithmic}
\end{algorithm}


Then, We define the importance-refined distribution and the effective sampling distribution at iteration $k$ as follows:
\begin{subequations}
\begin{align}
    p_{\text{imp}}^{(k)}(x)
&\propto
\mathbf{1}\!\left(x \in \mathcal{T}^{(k)}\right)
\, p_{\text{nom}}^{(k)}(x). \\
p_{\text{mix}}^{(k)}(x)
&=
(1-\beta)\, p_{\text{nom}}^{(k)}(x)
+
\beta\, p_{\text{imp}}^{(k)}(x),
\end{align}
\end{subequations}
where $\beta \in [0,1)$ balances global coverage and boundary refinement.

In practice, sampling from this high-dimensional mixture $p_{\text{mix}}^{(k)}$ is physically instantiated by dynamically adjusting the temporal stride along each newly collected nominal trajectory. Specifically, we construct the trajectory-specific index set $T_s^i$ (introduced in \cref{loss_eq}) by applying a dense temporal sampling stride to segments where states enter the importance region, and a sparse uniform stride elsewhere.

The aggregated training buffer is then updated as $\mathcal{D}^{(k+1)} = \mathcal{D}^{(k)} \cup \mathcal{D}_{\text{imp}}^{(k)}$. The monitor $\hat{V}_{\text{stop}}^{(k+1)}$ is subsequently retrained on $\mathcal{D}^{(k+1)}$. This iterative process progressively expands nominal coverage while concentrating sample density near the critical stoppability boundary, as summarized in Algorithm~\ref{alg:progressive_refine}.



\section{Experiments}

\begin{figure*}[t]
    \centering
    \includegraphics[width=\linewidth]{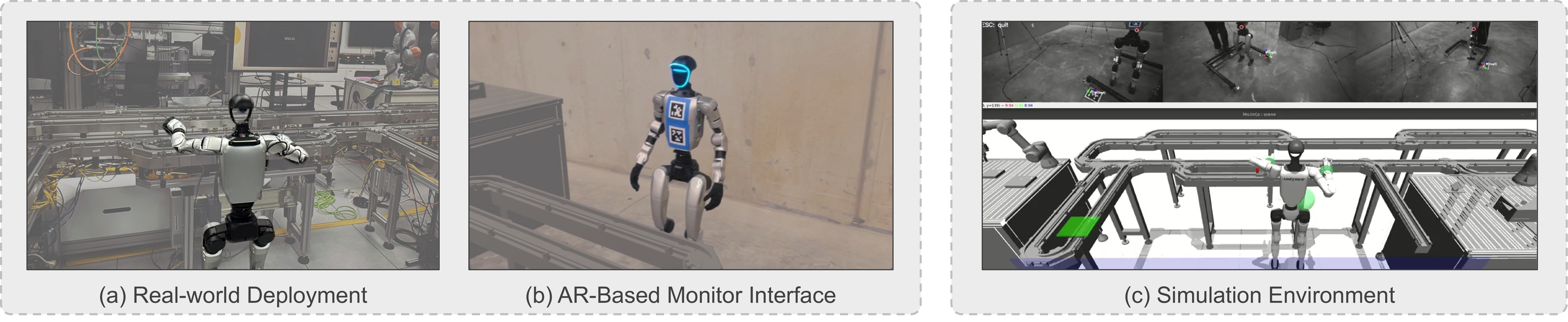} 
    \caption{
\textbf{Experimental platforms used}: 
(\textbf{Left}) Real humanoid deployment.
(\textbf{Middle}) AR based of data collection.
(\textbf{Right}) Simulation-based fallback policy rollouts. AR-based data collection improves safety during real-robot experiments by projecting the virtual environment onto the real robot, enabling realistic data collection while performing collision checking in simulation to avoid physical collisions.
}
    \label{fig:system_setup}
\end{figure*}
\subsection{System Setup. }
Experiments are conducted on the 29-DoF Unitree G1 humanoid performing a fixed loco-manipulation task: grasping an object from a MagneMotion Conveyor cart, navigating to a designated area, and releasing it. The nominal controller $\pi_{\text{nom}}$ is a pretrained whole-body policy executing in closed-loop via onboard proprioception and a three-camera global localization system. The safe set $\mathcal{X}_{\text{safe}}$ includes states where the robot is balanced, collision free, and meets joint-limit constraints. Its subset, the terminal MRC $\mathcal{X}_{\text{terminal}} \subset \mathcal{X}_{\text{safe}}$, designates a predefined upright stance with both arms in a default posture. We train the task policy $\pi_{\text{nom}}$ to perform loco-manipulation in the task area; and the task-agnostic fallback policy $\pi_{\text{fallback}}$ to drive the robot toward $\mathcal{X}_{\text{terminal}}$ from arbitrary initial states without considering environmental interaction (hence having no knowledge of the task-dependent $\mathcal{X}_{\text{safe}}$). 


\subsection{Data Collection. }
Stoppability labeling is conducted in the SPARK simulator~\cite{sun2025spark} using the identical nominal and fallback policies. During real-world trials, which are monitored and reconstructed via an Apple Vision Pro AR headset, we record proprioceptive and global camera states. These states are subsequently replayed in simulation to execute importance-guided E-stop rollouts. The rollouts yield binary labels: \texttt{Unsafe} if any collision occurs or the robot falls down, and \texttt{Safe} if the robot reaches the terminal set without incident. We collect 40 real-world trajectories and 400 simulated seeds, which are used to generate safety labels and to train the models for real-world and simulation experiments, respectively. For simulation evaluation, we use single rollouts for labeling, as \texttt{Unsafe} labels in our task typically occur when the lower body remains stationary while the upper body collides during the safe stop, resulting in quite deterministic labels even under domain randomization.

\subsection{Data-Efficient Iterative Monitor Learning}
In our importance sampling setting, the PRISM is initialized with $N_0=3$ trajectories, Then, $N_I = 3$ more novel trajectories are incorporated per iteration. Critical boundary regions identified by importance weights are sampled densely (\textit{fine stride}), whereas trivial safe regions are sampled sparsely (\textit{coarse stride}).

\begin{table}[t]
\centering
\caption{{Progressive Stoppability Monitor Refinement (Sim).}}
\label{tab:iteration}
\resizebox{\linewidth}{!}{%
\begin{tabular}{lcccccc}
\toprule
\multirow{2}{*}{\textbf{Sampling}} 
& \multirow{2}{*}{\textbf{Iter}} 
& \textbf{Total} 
& \textbf{Unsafe} 
& \textbf{Num.} 
& \multicolumn{2}{c}{\textbf{Pred Acc.} (\%) $\uparrow$} \\
\cline{6-7}
&  & \textbf{Data} 
& \textbf{Ratio} 
& \textbf{Traj.} 
& \textbf{Safe} 
& \textbf{Unsafe} \\
\midrule

\multicolumn{7}{l}{\cellcolor{gray!10}\textit{- Importance Sampling (PRISM)}} \\

\multirow{6}{*}{\centering PRISM}
& 0  & $715$  & $29.7\%$ & $3$  & $93.8$ & $70.0$ \\
& 1  & $1127$ & $28.4\%$ & $6$  & $92.7$ & $72.3$ \\
& 5  & $2754$ & $27.6\%$ & $18$ & $92.9$ & $74.2$ \\
& 9  & $4332$ & $26.9\%$ & $30$ & $93.2$ & $73.5$ \\
& 13 & $5801$ & $27.3\%$ & $42$ & $95.0$ & $77.2$ \\
& 17 & $7339$ & $27.4\%$ & $54$ & $94.3$ & $87.9$ \\

\multicolumn{7}{l}{\cellcolor{gray!10}\textit{- Full-Buffer Uniform Training}} \\

30 Traj. & -- & $7212$  & $28.2\%$ & $30$ & $96.4$ & $78.3$ \\
54 Traj. & -- & $12564$ & $28.7\%$ & $54$ & $94.8$ & $89.2$ \\

\bottomrule
\end{tabular}%
}
\end{table}

We benchmark PRISM against two full-buffer uniform training baselines (Table~\ref{tab:iteration}). The $30$-trajectory baseline matches the strict data volume of PRISM at Iteration 17, while the $54$-trajectory baseline matches PRISM's total trajectory exposure. In Table~\ref{tab:iteration}, under an equivalent data budget ($30$-Traj baseline), PRISM improves the prediction accuracy for critical unsafe states. Furthermore, PRISM achieves comparable performance with the $54$-trajectory baseline, yet reduces the total data collection footprint by over $40\%$. This clearly demonstrates PRISM's capability to disentangle monitor accuracy from sheer data volume.

\subsection{Qualitative Study}

\begin{figure}[t] 
    \centering
    \includegraphics[width=\linewidth]{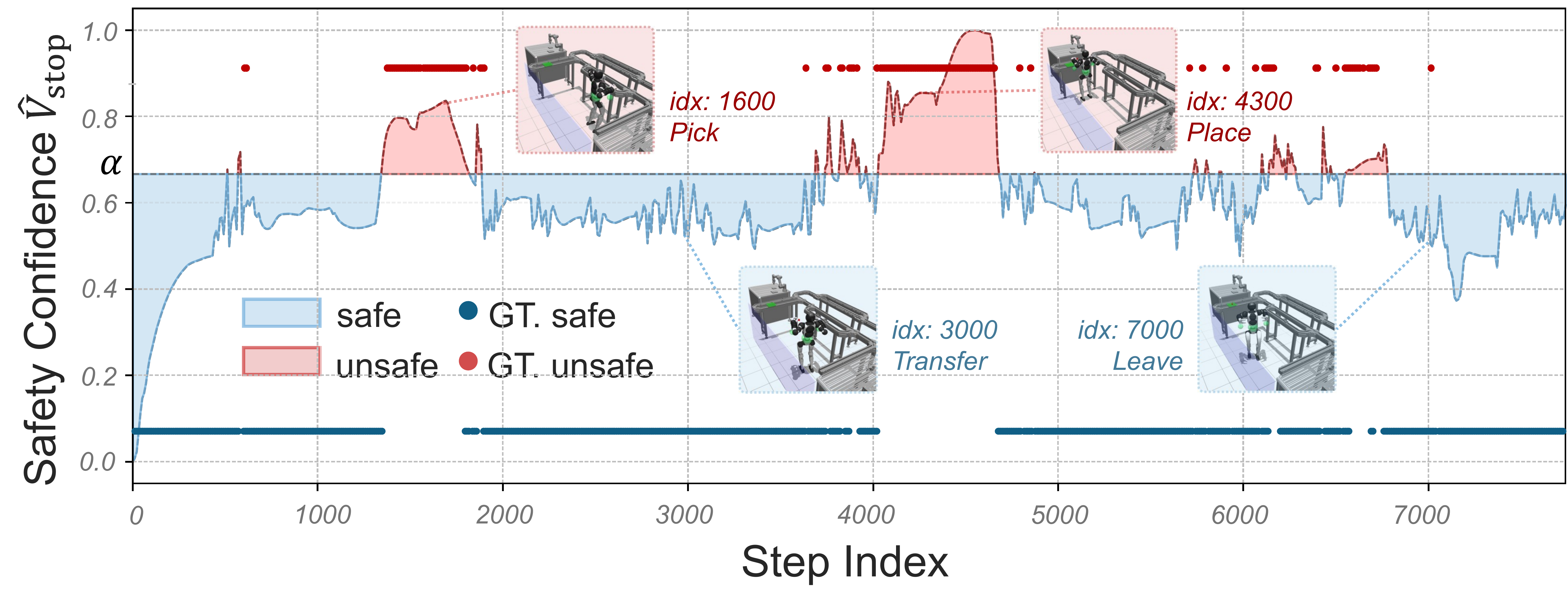} 
    \caption{\textbf{Safety score over a full loco-manipulation trajectory.} The monitor exhibits precise temporal alignment with the ground truth and identifies elevated risks during manipulation phases (\textit{Pick}, \textit{Place}) while ensuring high confidence during steady-state locomotion (\textit{Transfer}, \textit{Leave}).}
    \label{fig:safety_confidence}
\end{figure}

\begin{figure}[t]
    \centering
    \includegraphics[width=\linewidth]{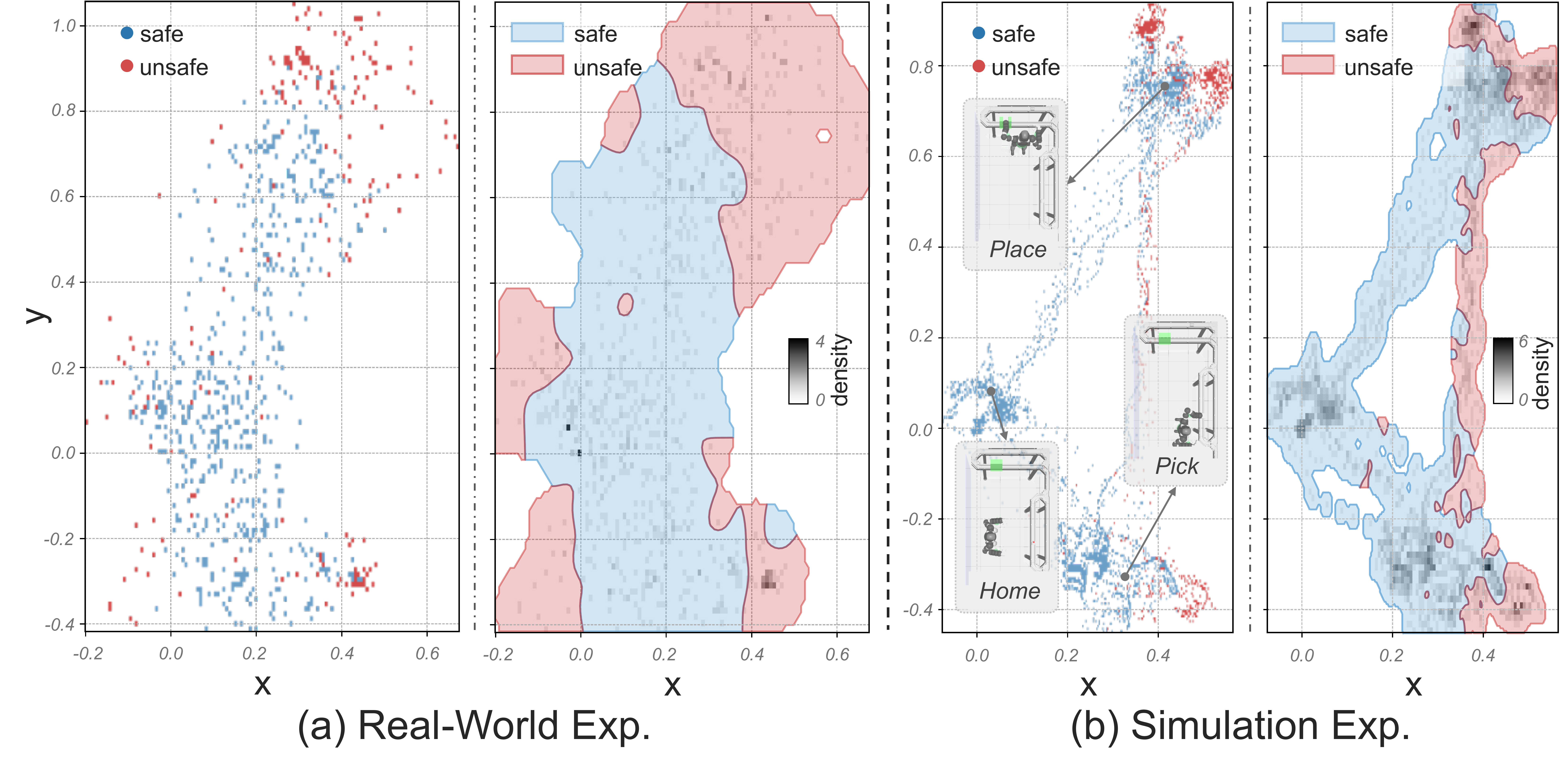} 
    \caption{\textbf{Spatial projection of the stoppability boundaries.} For both (a) Real-World and (b) Simulation, the scatter plots (left panels) display the Cartesian $(x,y)$ state distributions colored by ground-truth labels. The contour maps (right panels) illustrate the interpolated predictive safety regions overlaid with state density.}
    \label{fig:spatial_map}
\end{figure}

\noindent\textbf{Safety Confidence Profiling.} Figure~\ref{fig:safety_confidence} visualizes the continuous output logits of the trained stoppability monitor $\hat{V}_{\text{safe}}$ across a complete $\sim7500$-step nominal trajectory. During steady-state walking (e.g., \textit{Transfer}, \textit{Leave}), the confidence heavily fluctuates in the deep negative (\texttt{Safe}) region, reflecting the high stability margin of the nominal limit cycle. Conversely, severe probability spikes into the positive (\texttt{Unsafe}) domain correlate precisely with the \textit{Pick} and \textit{Place} phases. Our monitor successfully captures these complex, state-dependent dynamic vulnerabilities.

\noindent\textbf{Spatial State-Space Profiling.} Complementary to the temporal analysis, Figure~\ref{fig:spatial_map} visualizes the spatial distribution of the monitor's predictions using 2D Cartesian coordination $(x, y)$.  Crucially, the contour density maps demonstrate that the decision boundaries (the red/blue interfaces) in both the real-world deployment and the simulation environment exhibit highly consistent topological structures. This structural isomorphism strongly validates the monitor's sim-to-real transferability and its physical grounding in identifying spatially-induced failures.

\begin{table}[t]
\centering
\caption{{Single-Factor Domain Randomization Ablation (Sim).}}
\label{tab:dr_ablation}
\resizebox{\linewidth}{!}{%
\begin{tabular}{cccccc}
\toprule
\multirow{2}{*}{\textbf{DR ID}} 
& \multirow{2}{*}{\textbf{Perturbation}} 
& \multirow{2}{*}{\makecell{\textbf{Test}\\\textbf{Data}}}
& \multirow{2}{*}{\makecell{\textbf{Unsafe}\\\textbf{Ratio}}} 
& \multicolumn{2}{c}{\textbf{Pred Acc.} (\%) $\uparrow$} \\
\cline{5-6}
&  &  &  & \textbf{Safe} & \textbf{Unsafe} \\
\midrule

\multicolumn{6}{l}{\cellcolor{gray!10}\textit{- Default Configuration}} \\

Default & None & 8428 & 30.5 & 91.9 & 88.8 \\

\multicolumn{6}{l}{\cellcolor{gray!10}\textit{- Single-Factor Variations}} \\

0 & Damping $\downarrow$  & $7995$ & $30.1\%$ & $92.0$ & $88.2$ \\
1 & Damping $\uparrow$    & $7998$ & $30.1\%$ & $92.0$ & $88.2$ \\
2 & Gains $\downarrow$    & $7998$ & $30.2\%$ & $92.1$ & $88.0$ \\
3 & Gains $\uparrow$      & $8004$ & $30.3\%$ & $92.2$ & $88.0$ \\
4 & Friction $\downarrow$ & $8021$ & $30.7\%$ & $92.0$ & $86.5$ \\
5 & Friction $\uparrow$   & $8383$ & $36.8\%$ & $91.5$ & $74.7$ \\

\bottomrule
\end{tabular}%
}
\end{table}

\subsection{Robustness to Dynamics Perturbations}

Table~\ref{tab:dr_ablation} evaluates the zero-shot generalization of a single monitor (trained on $40$ nominal trajectories) against domain randomizations. Joint damping and geometry friction are randomized within $[0.5,1.6]$ and $[0.4,2.0]$ (default $[0.7,1.3]$). Motor gains are tested under $k_p\!\in\![0.75,1.25],\,k_d\!\in\![0.85,1.15]$ and $k_p\!\in\![0.6,1.4],\,k_d\!\in\![0.75,1.25]$, with default $k_p,k_d\!\in\![0.8,1.2]$. The monitor remains highly robust to internal joint parameters (damping and PD gains). Theoretically, these variations merely alter high-frequency transient control responses without fundamentally distorting the global whole-body kinematic envelope.

Conversely, the monitor is notably sensitive to external ground friction. Dynamically, excessively high friction prevents natural kinetic energy dissipation via micro-slipping. Consequently, the aggressive deceleration induced by $\pi_{\text{fallback}}$ abruptly converts linear momentum into angular tipping momentum around the foot edges. These novel tripping failure modes act as out-of-distribution (OOD) states.

\subsection{Sample Stride Sensitivity}

To investigate the inherent trade-off between temporal sampling resolution and data collection cost, we perform an ablation study on the trigger stride using a class-balanced dataset (Table~\ref{tab:stride_ablation}). As intuitively expected, decreasing the temporal stride yields denser state coverage, which steadily improves the overall prediction accuracy. Notably, the denser sampling also reduces false-safe errors by improving coverage near the stoppability boundary. However, this accuracy improvement incurs a steep empirical cost, proportionally inflating the expensive data collection time. This fundamental conflict between blanket dense sampling and hardware safety directly substantiates the necessity of our PRISM framework.

\begin{table}[t]
\centering
\caption{{Ablation Analysis of Stride Variation Sensitivity (Sim).}}
\label{tab:stride_ablation}
\resizebox{0.9\linewidth}{!}{%
\begin{tabular}{cccccc}
\toprule
\multirow{2}{*}{\textbf{Method ID}} 
& \multirow{2}{*}{\textbf{Stride}} 
& \multirow{2}{*}{\makecell{\textbf{Train}\\\textbf{Data}}}
& \multirow{2}{*}{\makecell{\textbf{Unsafe}\\\textbf{Ratio}}} 
& \multicolumn{2}{c}{\textbf{Pred Acc.} (\%) $\uparrow$} \\
\cline{5-6}
&  &  &  & \textbf{Safe} & \textbf{Unsafe} \\
\midrule
\multicolumn{6}{l}{\cellcolor{gray!10}\textit{- Stride Variations (Balanced Training)}} \\
0 & $60$ & $2624$ & $50.0\%$ & $91.00$ & $82.45$ \\
1 & $50$ & $3130$ & $50.0\%$ & $92.16$ & $81.79$ \\
2 & $40$ & $3866$ & $50.0\%$ & $91.57$ & $79.81$ \\
3 & $30$ & $5334$ & $50.0\%$ & $92.20$ & $84.63$ \\
4 & $20$ & $7864$ & $50.0\%$ & $\mathbf{92.35}$ & $\mathbf{84.55}$ \\

\bottomrule
\end{tabular}%
}
\end{table}

\subsection{Effect of Threshold Margin \texorpdfstring{$\alpha$}{alpha}}

To investigate the inherent trade-off between operational conservatism and task efficiency, we perform an ablation study on the decision threshold $\alpha$ (Table~\ref{tab:alpha_ablation}, we used stride-10 sampling and balanced the \texttt{Unsafe} and \texttt{Safe} labels.). As empirically demonstrated, tuning $\alpha$ directly governs the monitor's bias towards safety versus nominal utility. A lower threshold yields a highly conservative monitor, drastically elevating the \texttt{Unsafe} prediction accuracy to $99.73\%$. However, this extreme safety margin incurs a utility penalty.

\begin{table}[t]
\centering
\caption{{Ablation Analysis of Decision Threshold $\alpha$ (Sim).}}
\label{tab:alpha_ablation}
\resizebox{0.95\linewidth}{!}{%
\begin{tabular}{lcclcc}
\toprule
\multirow{2}{*}{\textbf{Param $\alpha$}} & \multicolumn{2}{c}{\textbf{Pred Acc.} (\%) $\uparrow$} & 
\multirow{2}{*}{\textbf{Param $\alpha$}} & \multicolumn{2}{c}{\textbf{Pred Acc.} (\%) $\uparrow$} \\
\cmidrule(lr){2-3} \cmidrule(lr){5-6}
& \textbf{Safe} & \textbf{Unsafe} & & \textbf{Safe} & \textbf{Unsafe} \\
\midrule
$0.47$ & $62.05$ & $99.38$ & $0.53$ & $93.58$ & $94.05$ \\
$0.49$ & $78.56$ & $98.44$ & $0.55$ & $96.82$ & $89.14$ \\
$0.51$ & $88.32$ & $96.46$ & $0.57$ & $98.86$ & $81.52$ \\
\bottomrule
\end{tabular}%
}
\end{table}

\subsection{Real-World Experiments}
To validate empirical efficacy and sample efficiency, we evaluate the trained monitor on the physical Unitree G1 humanoid using a hold-out test set of 5 unseen nominal trajectories. In safety-critical hardware deployments, the cost of a false negative exponentially outweighs a false positive. Notably, PRISM inherently learns a more conservative decision boundary, reflected by a lower \texttt{Safe} accuracy ($70.0\%$). Given the severe asymmetric penalty between false alarms and hardware destruction, this conservative prediction margin is highly desirable for physical robotic systems, definitively validating PRISM's capacity to maximize operational safety while minimizing hazardous real-world data collection.



\begin{table}[t]
\centering
\caption{{Progressive Stoppability Monitor Refinement (Real).}}
\label{tab:iteration-REAL}
\resizebox{\linewidth}{!}{%
\begin{tabular}{lcccccc}
\toprule
\multirow{2}{*}{\textbf{Sampling}} 
& \multirow{2}{*}{\textbf{Iter}} 
& \textbf{Total} 
& \textbf{Unsafe} 
& \textbf{Num.} 
& \multicolumn{2}{c}{\textbf{Pred Acc.} (\%) $\uparrow$} \\
\cline{6-7}
&  & \textbf{Data} 
& \textbf{Ratio} 
& \textbf{Traj.} 
& \textbf{Safe} 
& \textbf{Unsafe} \\
\midrule

\multicolumn{7}{l}{\cellcolor{gray!10}\textit{- Importance Sampling (PRISM)}} \\

PRISM & $11$ & $4099$ & $35.3\%$ & $33$ & $70.0$ & $\mathbf{94.1}$ \\

\multicolumn{7}{l}{\cellcolor{gray!10}\textit{- Full-Buffer Uniform Training}} \\

20 Traj. & -- & $4169$  & $34.9\%$ & $20$ & $82.2$ & $88.7$ \\
33 Traj.  & -- & $6820$ & $35.1\%$ & $33$ & $77.3$ & $93.3$ \\

\bottomrule
\end{tabular}%
}
\end{table}

\section{Discussion}


\noindent\textbf{Safety Integrity Level Analysis.} 
To assess deployment readiness against industrial benchmarks, we evaluate our framework's performance contextually within Safety Integrity Levels (SIL). By strategically tuning the decision threshold $\alpha$ to enforce a conservative safety bias, the monitor consistently elevates the prediction accuracy of critical \texttt{Unsafe} states above $99\%$ (e.g., see $\alpha=0.47$ in \cref{tab:alpha_ablation}). In the domain of functional safety, an automated hazard detection and intervention reliability exceeding $99\%$ fundamentally satisfies the stringent risk-reduction requirements of SIL 2. Consequently, our tunable safe-stoppability monitor can fullfill the rigorous prerequisites essential for the practical production and real-world deployment of bipedal humanoids.


\noindent\textbf{Sim-to-Real Considerations.}
Two sources of sim-to-real discrepancy arise: (1) the distribution gap of triggering states and (2) dynamics mismatch during fallback execution. The first is mitigated by collecting nominal trajectories on the real robot to initialize fallback rollouts in simulation, ensuring realistic triggering states. The second is mitigated by domain randomization during fallback rollouts in simulation. These enable robust approximation of the safe-stoppable envelope (SSE). Finally, the decision threshold $\alpha$ can be tuned to adjust the conservatism of the learned SSE, helping to compensate for residual sim-to-real mismatch introduced by domain randomization.

\noindent\textbf{Limitations.}
The approach depends on the fidelity of simulation and the representativeness of sampled scenarios. Conservative thresholding improves safety at the cost of frequent interventions. Future work includes integrating monitor outputs into runtime switching logic and extending the method to multi-stage emergency behaviors.

\section{Conclusion}
We presented a simulation-driven framework for fail-safe safety in humanoid robots by modeling safety as policy-dependent stoppability under a predefined safe-stop controller. We learn a neural monitor from large-scale simulation labels and introduced an iterative importance sampling strategy to refine performance in safety-critical failure regions. Experiments across humanoid scenarios demonstrate improved failure-case prediction and reduced false-safe errors under a fixed simulation budget. 

Beyond monitoring, the characterized safe-stoppable envelope naturally enables a closed-loop improvement process. Insights from identified failure modes and boundary states can be used to systematically refine the fallback controller, progressively enlarging the safe-stoppable, and therefore operational envelope. Such iterative co-design between monitoring and control holds significant promise for scalable industrial deployment, and will be explored in future work.
In addition, for states that currently lie outside the safe-stoppable envelop, one may design secondary recovery policies that first steer the system back into the envelope before executing the stop. This layered fallback architecture offers a principled pathway toward broader recoverability and enhanced runtime safety.

\section*{Acknowledgment}
This project was supported in part by Siemens and in part by the National Science Foundation under Grant No.~2144489.

\bibliographystyle{IEEEtran}
\bibliography{reference}
\end{document}